\documentclass{article}

\usepackage{arxiv}

\usepackage[utf8]{inputenc} 
\usepackage[T1]{fontenc}    
\usepackage{hyperref}       
\usepackage{url}            
\usepackage{booktabs}       
\usepackage{amsfonts}       
\usepackage{nicefrac}       
\usepackage{microtype}      
\usepackage{lipsum}		
\usepackage{multicol}
\usepackage{multirow}
\usepackage{graphicx}
\usepackage[numbers]{natbib}
\usepackage{doi}
\usepackage{algorithm}
\usepackage{algpseudocode} 
\usepackage{pifont}
\usepackage{amsmath}

\title{EXGnet: a single-lead explainable-AI guided multiresolution network with train-only quantitative features for trustworthy ECG arrhythmia classification}

\author{ 
{\includegraphics[scale=0.06]{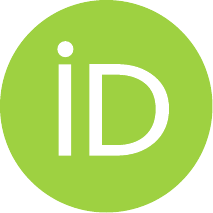}\hspace{1mm}Tushar Talukder Showrav} \\
	Dept. of Electrical and Electronic Engineering\\ Bangladesh University of Engineering and Technology\\ Dhaka, Bangladesh\\
	\texttt{1706087@eee.buet.ac.bd} \\
	\And
    {\includegraphics[scale=0.06]{orcid.pdf}\hspace{1mm}Soyabul Islam Lincoln} \\
	Dept. of Electronics \& Communication Engineering\\ Khulna University of Engineering and Technology\\ Khulna, Bangladesh\\
	\texttt{islam1709028@stud.kuet.ac.bd} \\
    \And
{\includegraphics[scale=0.06]{orcid.pdf}\hspace{1mm}Md Kamrul Hasan}\thanks{corresponding author} \\
	Dept. of Electrical and Electronic Engineering\\ Bangladesh University of Engineering and Technology\\ Dhaka, Bangladesh\\
	\texttt{khasan@eee.buet.ac.bd} \\
}

\renewcommand{\shorttitle}{Trustworthy ECG Arrhythmia Classification}

\begin{document}
\maketitle

\begin{abstract}
Deep learning has significantly propelled the performance of ECG arrhythmia classification, yet its clinical adoption remains hindered by challenges in interpretability and deployment on resource-constrained edge devices. To bridge this gap, we propose EXGnet, a novel and reliable ECG arrhythmia classification network tailored for single-lead signals, specifically designed to balance high accuracy, explainability, and edge compatibility. 
EXGnet integrates explainable artificial intelligence (XAI) supervision during training via a normalized cross-correlation based loss, directing the model’s attention to clinically relevant ECG regions, similar to a cardiologist’s focus. This supervision is driven by automatically generated ground truth, derived through an innovative heart rate variability-based approach, without the need for manual annotation.
To enhance classification accuracy without compromising deployment simplicity, we incorporate quantitative ECG features during training. These enrich the model with multi-domain knowledge but are excluded during inference, keeping the model lightweight for edge deployment. Additionally, we introduce an innovative multiresolution block to efficiently capture both short- and long-term signal features while maintaining computational efficiency.
Rigorous evaluation on the Chapman and Ningbo benchmark datasets validates the supremacy of EXGnet, which achieves average five-fold accuracies of 98.762\% and 96.932\%, and F1-scores of 97.910\% and 95.527\%, respectively. Comprehensive ablation studies and both quantitative and qualitative interpretability assessment confirm that the XAI guidance is pivotal, demonstrably enhancing the model's focus and trustworthiness.
Overall, EXGnet sets a new benchmark by combining high-performance arrhythmia classification with interpretability, paving the way for more trustworthy and accessible portable ECG based health monitoring systems.
\end{abstract}

\keywords{Single-lead ECG systems \and ECG arrhythmia classification \and Trustworthy \and Explainable Artificial Intelligence \and XAI guidance}

\section{Introduction}

Cardiovascular disease (CVD) is the leading cause of death globally, affecting over half a billion people, with 20.5 million deaths in 2021 \cite{who_cardio}. More than 80\% of these fatalities result from heart attacks and strokes, with one-third occurring prematurely in individuals under 70 years old \cite{ncd2020repositioning}. These CVDs can be detected and monitored using various diagnostic tools. One of the most effective tools is the electrocardiogram (ECG), a fast and simple test that measures heart rate, rhythm, and timing by recording natural electrical impulses in the heart's various parts \cite{abmabdullah}. Recent advances in deep learning and edge computing have led to cost-effective and user-friendly portable ECG devices, especially single-lead systems, which strongly support smartphone-based ECG analysis at home and empower individuals to monitor their heart health conveniently and continuously \cite{tang2022near}.

Within the evolving landscape, two main categories of CVD detection technologies have emerged: traditional methods and deep learning-based methods. Traditional methods comprise an expert feature extraction stage and a machine learning algorithm prediction stage. Support Vector Machine (SVM) classifiers are widely utilized among machine learning methods \cite{islam2023hardc}. Early research used Discrete Wavelet Transform (DWT) to extract QRS morphology, but its limitation to low-frequency decomposition led to Wavelet Packet Decomposition (WPD), which includes both high- and low-frequency components \cite{kutlu2012feature}. Recent work has demonstrated that combining Time Series Feature Extraction Library (TSFEL)-based feature extraction with balanced ensemble learning, specifically a weighted majority vote of Random Forest and SVM classifiers, can significantly improve arrhythmia classification accuracy on the MIT-BIH database \cite{9440796}. The main drawback of these traditional methods is that they require hand-crafted features that are dependent on proper ECG signal processing and human intervention. A low-quality or noisy signal fails to provide proper features effectively. Furthermore, experimenting with various feature reduction techniques is essential to identify the optimal set of features.

\begin{figure}[t]
\centering
\includegraphics[width=0.5\linewidth]{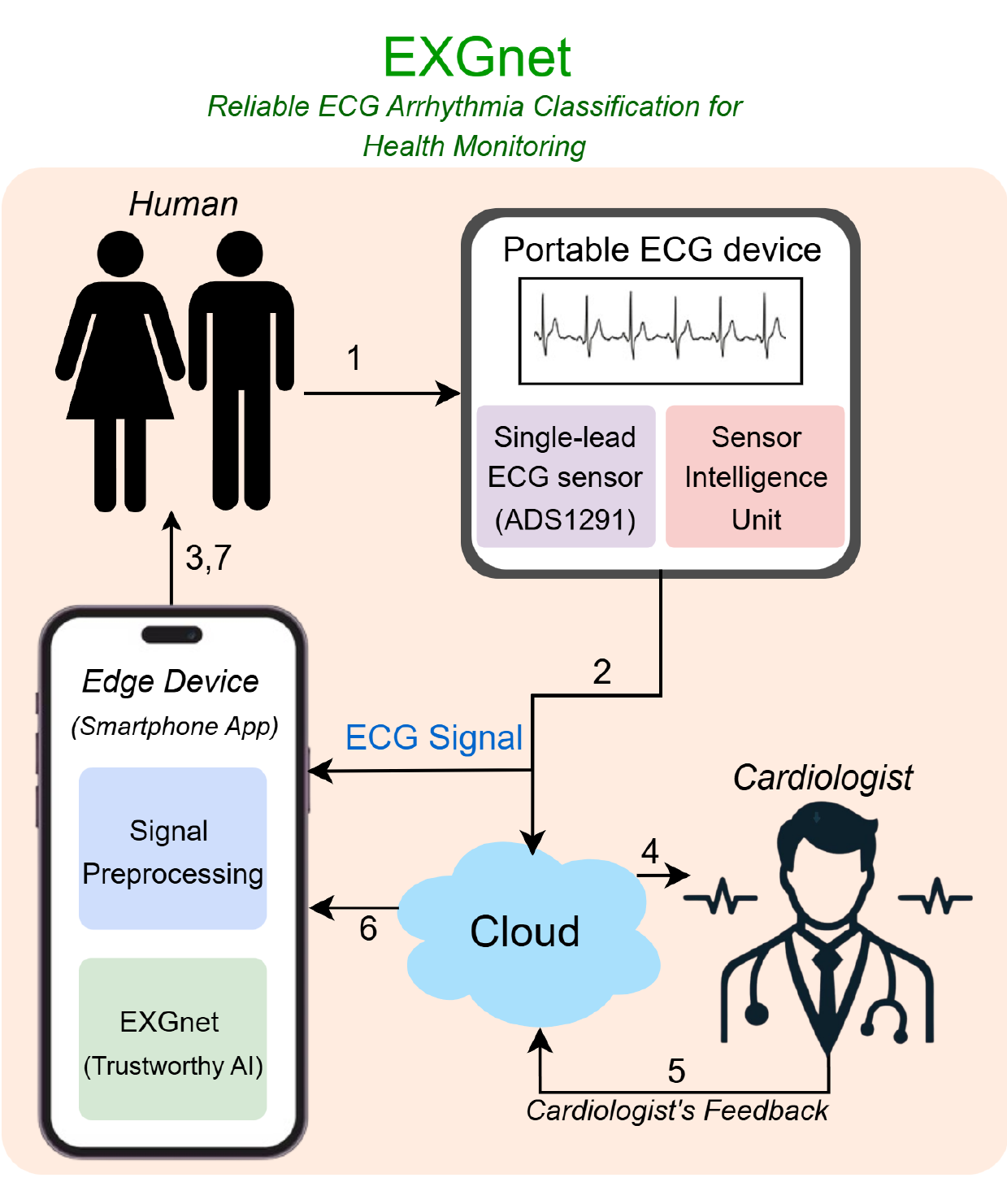}
\caption{Overview of ECG healthcare system where EXGnet can be integrated for reliable real-time heart arrhythmia detection.}
\label{fig:exg_overview}
\end{figure}

To overcome the limitations of traditional approaches, deep Convolutional Neural Networks (CNNs) have gained significant attention from researchers due to their powerful ability to automatically extract and select features directly from input signals. Asgharzadeh \emph{et al.} \cite{asgharzadeh2020spectral} proposed calculating spectral entropy to generate a three-dimensional time-frequency transform for QRS complexes and employed bi-directional, two-dimensional Principal Component Analysis (PCA) to filter out irrelevant features and enhance class separation by simultaneously applying PCA to the rows and columns of the Short-Time Fourier Transform (STFT), followed by a CNN for classification. Based on this, a novel hidden attention residual network was proposed to extract useful features from ECG signals by converting 2-second segments into images \cite{guan2022ha}. Additionally, a lightweight fusing transformer model was introduced to address challenges related to large-scale parameters and performance degradation, further advancing the capabilities of ECG analysis \cite{meng2022enhancing}. Tao \emph{et al.} \cite{tao2022ecg} incorporated an expert-knowledge attention network to recognize four tachyarrhythmias, highlighting the need for deep learning approaches to consider inter-channel shared information and time-sequence dependent signals. In another study \cite{bechinia2024approach}, a lightweight custom CNN and a transfer learning model with the MobileNet-V2 architecture were applied to segmented ECG image data that was balanced by an auxiliary classifier generative adversarial network. Chen \emph{et al.} \cite{chen2024emcnet} developed an ensemble multiscale CNN with two modules: a Long Short-Term Memory (LSTM) network that learns from 1-D signals and a module that learns from time-frequency spectrograms from denoised ECG signals using VGGNet. The Continuous Wavelet Transform (CWT) scalogram from a single-lead ECG signal was employed in a study by Xie \emph{et al.} \cite{xie2024automated}. Sarankumar \emph{et al.} \cite{sarankumar2024bidirectional} proposed a bidirectional Gated Recurrent Unit (GRU) with an autoencoder to detect ECG arrhythmias. A two-stage ECG classifier is proposed in \cite{xiaolin2024} that incorporates both ML models and CNNs. Zabihi \emph{et al.} \cite{zabihi2024electrocardiogram} proposed an innovative hybrid machine learning and deep learning approach to classify ECG beats. In a different study, spectrograms of 1D-ECG signals were created using the superlet-transform-based (SLT) method and classified with ResNet-18, as an alternative to the CWT \cite{singhal2024gsmd}. In another study, an innovative deep learning approach was proposed that incorporates a spectral correlation function to detect atrial fibrillation \cite{mihandoost2024deep}. Gupta \emph{et al.} proposed CardioNet, combining Short-Time Fourier Transform (STFT), Gaussian window-based S-transform (ST), and Smoothed Pseudo-Wigner–Ville Distribution (SPWVD), followed by a CNN for prediction \cite{10824221}. Islam \emph{et al.} \cite{islam2024cat} proposed an attention, convolution, and transformer-based network to effectively capture heartbeat morphological characteristics using peak annotation on ECG signals, incorporating preprocessing algorithms such as Synthetic Minority Oversampling Technique-Edited Nearest Neighbor (SMOTE-ENN), SMOTE-Tomek, and Adaptive Synthetic (ADASYN) sampling. A hybrid deep learning framework was proposed in \cite{guhdar2025advanced}, which incorporates a specialized attention mechanism. In another approach, a novel inter-patient heartbeat classification model, G2-ResNeXt, was proposed to enhance the recognition of patients' heart conditions by incorporating a two-fold grouping convolution (G2) into the original ResNeXt structure \cite{hao2023g2}. To tackle class imbalances in ECG data, a context-aware loss was introduced in \cite{el2024ecgtransform}, which dynamically adjusts weights according to class representation.

Explainable Artificial Intelligence (XAI) methods are relatively new in the field of deep learning-based biomedical signal analysis, with only a few studies exploring various XAI techniques for ECG classification. Jo \emph{et al.} \cite{jo2021explainable} developed a CNN architecture with submodules to detect the presence of P waves and RR irregularities, using Gradient-weighted Class Activation Mapping (Grad-CAM) to interpret the results. Bender \emph{et al.} \cite{bender2023analysis} employed multiple Layer-wise Relevance Propagation (LRP) rules and Integrated Gradients (IG) for interpreting different arrhythmias. In contrast, Singh \& Sharma \cite{singh2022interpretation} performed a more comprehensive comparison of four XAI methods, including Grad-CAM and Shapley Additive Explanations (SHAP). Goettling \emph{et al.} \cite{goettling2024xecgarch} compared 13 different XAI methods, and among those techniques, the Deep Taylor Decomposition (DTD) explanation was identified as the most trustworthy. H. Le \emph{et al.} \cite{le2023lightx3ecg} used modified Grad-CAM for lead-wise explanation. They employed Spearman's rank correlation to assess the quality of the XAI approach. In another work, Ayano \emph{et al.} \cite{ayano2024interpretable} employed two model interpretability techniques, namely: Grad-CAM++ and SHAP.

Most deep learning studies, whether based on CNNs, Recurrent Neural Networks (RNNs), or a hybrid of both, usually train models using either raw ECG signals or hand-crafted features. However, some arrhythmias have identifiable features that appear as specific patterns in the signal shape; therefore, accurate interpretation requires analyzing both the signal and its morphology. A few works have merged quantitative features with deep networks and reported performance gains, but this dependence on quantitative feature inputs complicates real-time deployment on resource-constrained edge devices such as smartphones. At the same time, XAI tools in the literature are used mainly as post-hoc demonstrations of a network’s trustworthiness; they seldom feed back into the learning process itself to foster more reliable predictions. In this paper, these limitations are addressed by introducing an innovative architecture, EXGnet, which provides a reliable, trustworthy, and interpretable deep learning framework for ECG analysis. The major contributions of this paper are as follows:
\begin{enumerate}
\item An innovative deep learning architecture is proposed that efficiently extracts multiresolution features from the ECG signal while minimizing computational complexity. This approach enables the model to learn both short-term and long-term features within a single network.

\item XAI guidance during training is introduced through a Normalized Cross-Correlation (NCC)-based loss function to guide Class Activation Map (CAM) values. This strategic approach ensures that the proposed deep learning model focuses on the specific regions of ECG signals that cardiologists typically examine during their decision-making.

\item The ground truth CAMs for XAI guidance are automatically generated using an innovative heart rate variability-based method without the need for manual annotation.

\item The ECG quantitative features are strategically integrated solely during the training phase, allowing the network to incorporate them for enhanced learning. Remarkably, during testing, these features are not required, yet the model demonstrates improved performance compared to when they were excluded during training.

\item The innovative quantitative approach for interpretability analysis, combined with qualitative visualizations, highlights the reliability and trustworthiness of the proposed network.

\end{enumerate}

By bringing together these innovative components in a single framework, we offer a strong and flexible solution that works well in both real-world and remote settings. Figure~\ref{fig:exg_overview} shows where EXGnet can be implemented in a portable health monitoring system. We evaluated our model on two publicly available and widely-used ECG arrhythmia datasets, Chapman and Ningbo, and the results show that it performs better than recent state-of-the-art methods by a clear margin. The inclusion of XAI guidance during training further improves the trustworthiness of the network by making its decisions more understandable and reliable. The rest of this paper is arranged as follows. Section~\ref{materials_methods} describes the materials and methods, including information about the datasets, how we created the ground truth CAMs, the network design, and the loss functions. Section~\ref{results} presents the experimental setup and comparative results. Section~\ref{discussion} includes an ablation study, an analysis of the model’s reliability, confusing cases identified in the Ningbo dataset, and discusses possible limitations. Lastly, Section~\ref{conclusion} summarizes our findings.

\begin{figure*}[t]
\centering
\includegraphics[width=\linewidth]{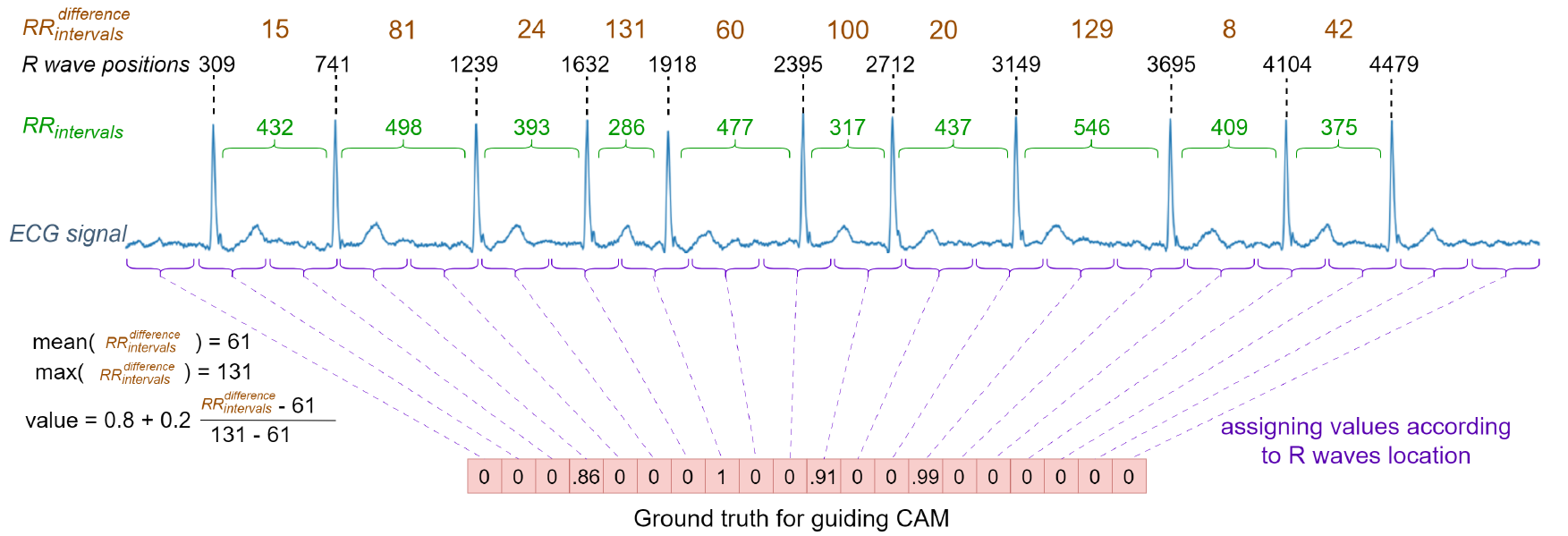}
\caption{Illustration of ground truth generation for guiding CAM with a sample ECG signal. The presented ECG signal consists of 5000 samples, corresponding to a 10-second duration at a sampling rate of 500 Hz. All measurements in the figure are presented in terms of samples.}
\label{fig:cam}
\end{figure*}

\section{Materials and methods}
\label{materials_methods}
\subsection{Materials}
\subsubsection{Datasets}
To evaluate the performance of EXGnet, all experiments are conducted using the publicly available database from the PhysioNet/Computing in Cardiology Challenge 2021 \cite{reyna2021will}. This database includes ECG recordings from various challenges and hospitals, totaling 45,512 recordings from the Chapman-Shaoxing \cite{zheng202012} (Chapman dataset) and Ningbo \cite{zheng2020optimal} datasets, each sampled at 500 Hz for 10 seconds. The rationale behind selecting these two datasets is their widespread use in the existing literature.

The following heart conditions are focused on in this research: Sinus Rhythm (SR), Sinus Arrhythmia (SA), Sinus Tachycardia (ST), Sinus Bradycardia (SB), Atrial Fibrillation (AFib), Atrial Flutter (AF), Atrial Tachycardia (AT), and Supraventricular Tachycardia (SVT). These were selected for effective identification using the lead II signals. To enable multiclass classification, it is ensured that each ECG recording has a unique class, although some signals may belong to multiple categories. This resulted in 5,086 recordings from the Chapman dataset and 16,411 from the Ningbo dataset. Due to insufficient data, AF with AFib and AT with SVT are combined. In addition, the existing literature shows that the AF and AFib signals are often quite similar in these datasets \cite{ribeiro2023can}. The class distribution is detailed in Table \ref{table_dataset}. Due to the limited availability of only 15 samples for the combined AT and SVT class in the Ningbo dataset, this class was excluded from the analysis.

\begin{table}[ht]
\small
\centering
\setlength\tabcolsep{3pt}
\caption{Class distribution across different datasets}
\begin{tabular}{lcccccc}
\hline
\multirow{2}{*}{Database} & \multicolumn{6}{c}{Class}                    \\ 
                          & SR   & SB   & ST   & SA   & AF+AFib & AT+SVT \\\hline
Chapman          & 1362 & 2195 & 639  & 0    & 501     & 390    \\
Ningbo                    & 4535 & 6691 & 2563 & 1232 & 1375    & 15     \\ \hline
\end{tabular}\label{table_dataset}
\end{table}

\subsubsection{Quantitative features extraction}
In our work, we utilized 17 heart rate variability based quantitative features extracted from ECG signals using the NeuroKit 0.2 Python library \citep{Makowski2021neurokit}. This library offers a comprehensive set of functions that can be applied to ECG data to extract essential features, aiding in detailed signal analysis. The following 17 features have been extracted from the ECG signal:  beats per minute (BPM), mean of normal-to-normal intervals (Mean NN), standard deviation of successive differences (SDSD), standard deviation of NN intervals (SDNN), and root mean square of successive differences (RMSSD). Additional features include the coefficient of variation of successive differences (CVSD), coefficient of variation of NN intervals (CVNN), median NN intervals (Median NN), median absolute deviation of NN intervals (MAD NN), and the median coefficient of variation of NN intervals (MCV NN). We also extracted the interquartile range of NN intervals (IQR NN), the standard deviation of the RMSSD (SDRMSSD), 20th and 80th percentiles of NN intervals (Prc20 NN, Prc80 NN), minimum NN intervals (Min NN), maximum NN intervals (Max NN), and the heart rate turbulence index (HTI).

\subsubsection{Data preprocessing}
For preprocessing, we minimized interventions to preserve raw signal characteristics. We applied three median filters with widths of 0.2, 0.6, and 1.2 seconds to the input signal, then subtracted the filtered components from the original signal to emphasize high-frequency details. After denoising, we normalized the signal values to a range of 0 to 1.

To generate the ground truth mask for XAI guidance, we concentrated on RR interval variability. We first calculated all the RR intervals from the ECG signal and then determined the absolute mean difference from the mean RR interval. Using Equation \ref{eq5}, we computed values only when $RR_{intervals}^{difference}$ is greater than or equal to $mean(RR_{intervals}^{difference})$; otherwise, the values were set to 0.
\begin{equation}
    RR_{intervals}^{difference} = RR_{intervals} - RR_{intervals}^{mean}
    \label{eq4}
\end{equation}
\begin{equation} 
value = \alpha + (1-\alpha) \frac{RR_{intervals}^{difference} - mean(RR_{intervals}^{difference})}{max(RR_{intervals}^{difference}) - mean(RR_{intervals}^{difference}})
\label{eq5}
\end{equation}\\
In our work, through experimentation, we set $\alpha$ to 0.8. Since the last convolutional layer of our proposed network has a length of 20, we divided the 5000 samples into 20 segments, with each segment representing 250 samples of the original ECG signal. We then identified the R wave location (R wave position + half of the corresponding RR interval) in each segment. If a segment contained a single R wave, it received the corresponding value. If there were two R waves, the maximum value of those R waves was assigned to the segment. If no R waves were present in a segment, the value was set to 0. This process allowed us to generate the ground truth mask effectively. Figure \ref{fig:cam} shows ground truth generation for XAI guidance for a sample ECG signal.

\begin{figure*}[t]
\centering
\includegraphics[width=\linewidth]{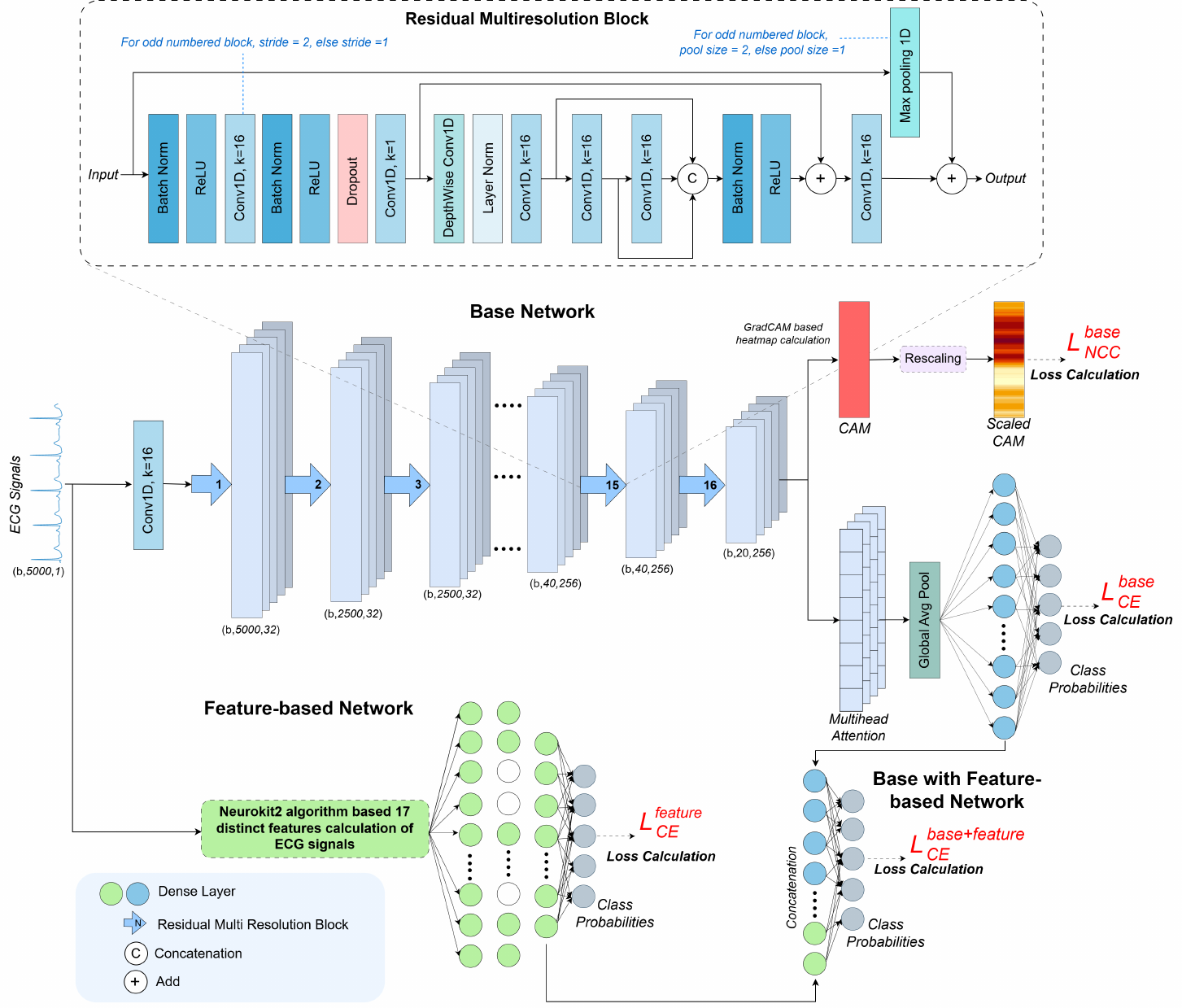}
\caption{Detailed block diagram of the proposed EXGnet.}
\label{fig:main}
\end{figure*}

\subsection{Proposed Network Architecture}
The proposed framework's architectural representation is depicted in figure \ref{fig:main}. This framework includes XAI guidance, features used exclusively during training, and an innovative base architecture including multiresolution block. The architecture of the framework, along with the motivation behind each design choice, is explained in detail in the following sections.

\subsubsection{Base network with multiresolution feature extraction}
Our proposed network takes a 10-second ECG signal as input, capturing both short-term and long-term features inherent in the signal. Figure~\ref{fig:main} shows the detailed block diagram of the proposed method. Unlike most existing approaches that focus exclusively on either short-term or long-term features, or employ separate CNN networks to extract and integrate both \cite{goettling2024xecgarch}, the proposed design takes a better approach. We introduce a residual multiresolution block that utilizes convolutions with varying kernel sizes and depth-wise convolutions to capture multiresolution features without the need for separate CNNs. To address the computational overhead associated with larger kernels, we draw inspiration from Navid \emph{et.al.} and Showrav \emph{et.al.} \cite{ibtehaz2020multiresunet,showrav2024hi}, implementing a parallel convolutional structure. Here, three 16$\times$16 kernel convolutions are used to get the advantage of both 16$\times$16, 31$\times$31, and 46$\times$46 convolutions, thus reducing the computation cost. This design achieves multiresolution feature extraction with reduced computational complexity. This process can be expressed as
\begin{equation}
  y = \overbrace{\psi^{16}(x)}^a \parallel \rlap{$\overbrace{\phantom{\psi^{16}(\psi^{16}(x))}}^b$}\psi^{16}(
     \underbrace{\psi^{16}(x)}_a) \parallel \rlap{$\overbrace{\phantom{\psi^{16}(\psi^{16}(\psi^{16}(x)))}}^c$}\psi^{16}(
     \underbrace{\psi^{16}(\psi^{16}(x))}_b)
\end{equation}
where $\psi^{k}$ denotes a convolutional layer with kernel size $k$, and $\parallel$ denotes concatenation. The residual multiresolution block also includes depth-wise convolution, enabling the network to learn features from individual channels. Similar to the work of Hannun \emph{et.al.} \cite{hannun2019cardiologist}, shortcut connections are incorporated to facilitate efficient optimization. The base network comprises 16 residual multiresolution blocks, with every alternate block subsampling its inputs by a factor of two. Corresponding Max Pooling layers within the blocks also perform subsampling on alternate layers. Batch normalization is applied before each ReLU activation layer, and a Dropout layer is used to mitigate overfitting. The number of channels per multiresolution block begins with 32 and doubles after every four blocks, ending with 256 in the final block.

Following the final residual multiresolution block, Multi-Head Attention is applied with 8 heads and an embedded dimension of 256, followed by a Global Average Pooling (GAP) layer. The network concludes with a Dense layer, the size of which corresponds to the number of arrhythmia classes, using Softmax activation to output the predicted class.

\subsubsection{XAI guidance}
In deep learning applications for arrhythmia detection, it is crucial to consider not only whether a model accurately predicts the condition but also how we can trust these predictions. Specifically, we need to determine if the results stem from the same regions of ECG signals that cardiologists traditionally rely on for their diagnoses. To address these concerns, XAI techniques can be employed to evaluate the trustworthiness and reliability of the model's output. Several studies in the literature have used methods such as Grad-CAM and Shapley values to interpret the decisions of neural networks, thereby enhancing confidence in their predictions \cite{goettling2024xecgarch}. However, to the best of our knowledge, no prior studies has investigated the integration of XAI guidance during the training phase.

In this paper, we introduce an innovative approach that incorporates XAI guidance directly into the training process. Our method integrates a CAM generation mechanism, inspired by Grad-CAM \cite{selvaraju2017grad}, within the network itself. Specifically, CAMs are produced from the output of the final multiresolution feature block, allowing the model to highlight signal regions that contribute most significantly to its predictions. The complete algorithm for CAM generation is outlined in Algorithm \ref{alg:ag_cam}. In our model, the CAM consists of 20 values, each corresponding to a sequential block of 250 samples from the original 5000-sample ECG signal. A CAM value close to 1 indicates that the corresponding region contains features critical to the final prediction. Previous research has shown that many severe heart conditions are associated with significant variability in the RR interval \cite{peltola2012role}. Therefore, we initially focus on RR interval variability in the ECG signal. To guide the network, we apply a 1D mask of 20 samples long generated by considering the RR intervals provided by the NeuroKit2 algorithm, which is designed to capture heart rate variability. By guiding the network with these masks, the model learns to concentrate on clinically relevant regions of the signal, thereby emulating the decision-making process of a cardiologist.

\begin{algorithm}[h]
\caption{Network-integrated CAM generation}
\label{alg:ag_cam}
\begin{algorithmic}[1]
\Statex \textbf{Input:} Model $M$, Input Signal $X$, Target Feature Layer $L$
\Statex \textbf{Output:} Class Activation Map (CAM) $H$
\Statex
\Procedure{GenerateCAM}{$M, X, L$}
    \State $A, Y_{scores} \gets \text{ForwardPass}(M, X, L)$
    \State $c \gets \arg\max(Y_{scores})$
    \State $y^c \gets Y_{scores}[c]$
    \State $G \gets \frac{\partial y^c}{\partial A}$ \Comment{Compute gradients w.r.t. feature map}
    \State $P \gets A \odot G$ \Comment{Element-wise product}
    \State $H_{raw} \gets \frac{1}{N_{channels}} \sum_{k} P_k$ \Comment{Average across channels}
    \State $H \gets \frac{H_{raw} - \min(H_{raw})}{\max(H_{raw}) - \min(H_{raw})}$ \Comment{Normalize to [0, 1]}
    \State \textbf{return} $H$
\EndProcedure
\end{algorithmic}
\end{algorithm}

\subsubsection{Train-only quantitative features integration}
Machine learning approaches to ECG arrhythmia classification rely on the extraction of various statistical, morphological, and handcrafted features from the ECG signal, followed by the utilization of classifiers such as Random Forest, Gradient Boosting, and Naive Bayes. However, these methods frequently fall short of providing optimal results. To improve performance, some studies have integrated these quantitative features with deep learning networks, yielding better outcomes. Despite these advances, the complexity of calculating numerous features poses a significant challenge for edge ECG devices, where computational efficiency is crucial.

To address this limitation, we introduce the use of train-only quantitative features, demonstrating that superior performance can still be achieved even when these features are excluded during inference. During testing, only the raw ECG signal is required, thereby eliminating the computational overhead of feature extraction and enhancing the system’s suitability for edge ECG devices. Specifically, we extract 17 distinct quantitative features, which are processed through a compact auxiliary network (Feature-based network) composed of three fully connected (dense) layers. The final layer of this network corresponds to the number of arrhythmia classes and employs a Softmax activation function. The output from the second dense layer is concatenated with the output of the GAP layer from the base ECG network. This fused representation is subsequently passed through an additional dense layer with Softmax activation to generate the final prediction.

The overall process can be expressed as:
\begin{equation}
y_j = J\left(\overline{G(x)}, \overline{F(x_f)}\right),
\end{equation}
where \( G(\cdot) \) denotes the base network operating on the raw ECG input \( x \), \( F(\cdot) \) represents the feature network processing the quantitative feature vector \( x_f \), and \( J(\cdot) \) denotes the joint network responsible for final classification. Here, \(\overline{G(x)}\) and \(\overline{F(x_f)}\) represent intermediate outputs of the base and auxiliary networks, respectively, which are fused for the final prediction. Although the primary emphasis of our approach lies in the base network, the quantitative features are not available during testing. However, the strategic incorporation of these train-only features during training enables the base network to learn enriched representations. This is achieved through a reciprocal loss mechanism, ultimately improving the overall classification performance.

\subsection{Loss functions}
We used Categorical Cross-Entropy (CE) loss during training of the network to learn arrhythmia labels. This can be expressed as
\begin{equation}
L_{CE} = - \sum_{i=1}^{N} \sum_{c=1}^{C} y_{i,c} \log(\hat{y}_{i,c})
\end{equation}
where, $N$ is the number of samples, $C$ is the number of classes, and $\hat{y}_{i,c}$ is the predicted probability where the ground truth probability is ${y}_{i,c}$. Besides, to guide the network with XAI, we use a network-integrated CAM generation technique. To guide the CAM values we utilize NCC loss function, which helps the network to learn making the CAM values orientation as per the ground truth. The NCC loss can be expressed as
\begin{equation}
L_{NCC} = - \frac{1}{N} \sum_{i=1}^{N} \frac{\sum_{j=1}^{M} \left( \frac{A_{i,j} - \mu_A}{\sigma_A} \right) \left( \frac{B_{i,j} - \mu_B}{\sigma_B} \right)}{M}\end{equation}
where, $N$ is the number of samples, $M$ is the number of values in CAM, and $A$ and $B$ denote the ground truth and prediction, respectively. The total loss function can be written as
\begin{equation}
L_{total} = \alpha \cdot L_{CE}^{base} + \beta \cdot L_{CE}^{feature} + \gamma \cdot L_{CE}^{base+feature} + \delta \cdot L_{NCC}^{base}
\end{equation}
where, $L_{CE}^{base}$ is the loss function for base network, $L_{CE}^{feature}$ for guiding the train-only quantitative feature based network, $L_{CE}^{base+feature}$ for guiding both base with feature based network, and $L_{NCC}^{base}$ for guiding the CAM values in the base network. In our work, through experimentation we selected $\alpha, \beta, \gamma$ and $\delta$ values as 2, 1, 1, and 0.2, respectively.

\begin{table*}[t]
\caption{Quantitative comparison of the proposed network with most recent state-of-the-art methods.}
\centering
\small
\setlength\tabcolsep{2.8pt}
\begin{tabular}{lccccccccc}
\hline
\multicolumn{1}{c}{\multirow{2}{*}{\textbf{Methods}}}                                                                                                              & \multirow{2}{*}{\textbf{Folds}}       & \multicolumn{4}{c}{\textbf{Chapman}}                                                       & \multicolumn{4}{c}{\textbf{Ningbo}}                                                 \\
\multicolumn{1}{c}{}                                                                                                                                               &                                       & \textit{Accuracy} & \textit{F1-score} & \textit{Sensitivity} & \textit{Specificity}        & \textit{Accuracy} & \textit{F1-score} & \textit{Sensitivity} & \textit{Specificity} \\ \hline
\multirow{6}{*}{\begin{tabular}[c]{@{}l@{}}Interpretable Hybrid\\ Multichannel Deep \\ Learning Model \citep{ayano2024interpretable}\end{tabular}} & \multicolumn{1}{c|}{\textit{Fold 1}}  & 95.481            & 93.654            & 94.248               & \multicolumn{1}{c|}{98.873} & 93.049            & 87.383            & 85.630               & 98.074               \\
     & \multicolumn{1}{c|}{\textit{Fold 2}}  & 95.088            & 92.657            & 92.300               & \multicolumn{1}{c|}{98.759} & 89.970            & 83.024            & 81.237               & 97.240               \\
      & \multicolumn{1}{c|}{\textit{Fold 3}}  & 95.379            & 92.632            & 93.186               & \multicolumn{1}{c|}{98.814} & 91.552            & 87.227            & 85.886               & 97.752               \\
    & \multicolumn{1}{c|}{\textit{Fold 4}}  & 96.853            & 95.336            & 95.724               & \multicolumn{1}{c|}{99.204} & 92.925            & 86.745            & 85.591               & 98.101               \\
    & \multicolumn{1}{c|}{\textit{Fold 5}}  & 96.264            & 93.991            & 93.516               & \multicolumn{1}{c|}{99.047} & 89.814            & 84.025            & 83.685               & 97.279               \\
    & \multicolumn{1}{c|}{\textit{Average}} & 95.813            & 93.654            & 93.795               & \multicolumn{1}{c|}{98.940} & 91.462            & 85.681            & 84.406               & 97.690               \\ \hline
\multirow{6}{*}{G2-ResNeXt \citep{hao2023g2}}                                                                                                      & \multicolumn{1}{c|}{\textit{Fold 1}}  & 98.428            & 97.189            & 97.237               & \multicolumn{1}{c|}{99.619} & 96.951            & 94.545            & 95.209               & 99.206               \\
    & \multicolumn{1}{c|}{\textit{Fold 2}}  & 97.937            & 96.342            & 96.789               & \multicolumn{1}{c|}{99.475} & 96.248            & 94.895            & 94.567               & 99.027               \\
    & \multicolumn{1}{c|}{\textit{Fold 3}}  & 98.132            & 96.799            & 96.856               & \multicolumn{1}{c|}{99.544} & 96.279            & 94.392            & 93.479               & 99.023               \\
    & \multicolumn{1}{c|}{\textit{Fold 4}}  & 98.820            & 98.074            & 97.977               & \multicolumn{1}{c|}{99.700} & 96.920            & 95.301            & 94.351               & 99.173               \\
    & \multicolumn{1}{c|}{\textit{Fold 5}}  & 98.623            & 97.947            & 97.834               & \multicolumn{1}{c|}{99.631} & 97.011            & 95.535            & 94.772               & 99.207               \\
    & \multicolumn{1}{c|}{\textit{Average}} & 98.388            & 97.270            & 97.339               & \multicolumn{1}{c|}{99.594} & 96.682            & 94.934            & 94.476               & 99.127               \\ \hline
\multirow{6}{*}{xECGArch \cite{goettling2024xecgarch}}                                                                                            & \multicolumn{1}{c|}{\textit{Fold 1}}  & 98.723            & 97.955            & 97.890               & \multicolumn{1}{c|}{99.676} & 96.646            & 95.030            & 95.736               & 99.165               \\
    & \multicolumn{1}{c|}{\textit{Fold 2}}  & 98.232            & 96.632            & 96.413               & \multicolumn{1}{c|}{99.560} & 96.340            & 95.036            & 94.715               & 99.039               \\
    & \multicolumn{1}{c|}{\textit{Fold 3}}  & 98.525            & 97.328            & 97.034               & \multicolumn{1}{c|}{99.630} & 96.493            & 94.751            & 94.394               & 99.070               \\
    & \multicolumn{1}{c|}{\textit{Fold 4}}  & 98.722            & 97.841            & 97.696               & \multicolumn{1}{c|}{99.678} & 96.950            & 95.223            & 95.223               & 99.219               \\                                                                                                                   & \multicolumn{1}{c|}{\textit{Fold 5}}  & 98.623            & 97.499            & 97.275               & \multicolumn{1}{c|}{99.656} & 96.706            & 94.865            & 93.968               & 99.117               \\                                                                                        & \multicolumn{1}{c|}{\textit{Average}} & 98.565            & 97.451            & 97.262               & \multicolumn{1}{c|}{99.640} & 96.627            & 94.981            & 94.807               & 99.122               \\ \hline
\multirow{6}{*}{EXGnet (Ours)}                                                                                                                                    & \multicolumn{1}{c|}{\textit{Fold 1}}  & 99.116            & 98.542            & 98.430               & \multicolumn{1}{c|}{99.825} & 97.043            & 95.661            & 96.023               & 99.260               \\                                                                                                                      & \multicolumn{1}{c|}{\textit{Fold 2}}  & 98.428            & 97.043            & 97.019               & \multicolumn{1}{c|}{99.601} & 96.523            & 95.082            & 94.651               & 99.096               \\                                                                                                                      & \multicolumn{1}{c|}{\textit{Fold 3}}  & 98.427            & 97.452            & 97.487               & \multicolumn{1}{c|}{99.605} & 96.859            & 95.276            & 94.802               & 99.174               \\                                                                                                                        & \multicolumn{1}{c|}{\textit{Fold 4}}  & 98.820            & 98.069            & 98.055               & \multicolumn{1}{c|}{99.716} & 96.920            & 95.531            & 95.354               & 99.198               \\                                                                                                                         & \multicolumn{1}{c|}{\textit{Fold 5}}  & 99.017            & 98.443            & 98.257               & \multicolumn{1}{c|}{99.743} & 97.316            & 96.085            & 96.185               & 99.324               \\                                                                                                                         & \multicolumn{1}{c|}{\textit{Average}} & 98.762            & 97.910            & 97.850               & \multicolumn{1}{c|}{99.698} & 96.932            & 95.527            & 95.403               & 99.210               \\ \hline
\multirow{6}{*}{EXGnet-L (Ours)}                                                                                                                                  & \multicolumn{1}{c|}{\textit{Fold 1}}  & 99.312            & 98.764            & 98.617               & \multicolumn{1}{c|}{99.904} & 97.348            & 96.103            & 96.327               & 99.331               \\                                                                                                                         & \multicolumn{1}{c|}{\textit{Fold 2}}  & 98.722            & 97.974            & 97.903               & \multicolumn{1}{c|}{99.678} & 97.011            & 95.875            & 95.877               & 99.239               \\                                                                                                                    & \multicolumn{1}{c|}{\textit{Fold 3}}  & 98.623            & 97.759            & 97.603               & \multicolumn{1}{c|}{99.648} & 97.133            & 95.825            & 95.887               & 99.264               \\                                                                                                                       & \multicolumn{1}{c|}{\textit{Fold 4}}  & 98.722            & 97.850            & 97.764               & \multicolumn{1}{c|}{99.682} & 97.225            & 96.140            & 96.109               & 99.286               \\                                                                                                                        & \multicolumn{1}{c|}{\textit{Fold 5}}  & 99.017            & 98.628            & 98.390               & \multicolumn{1}{c|}{99.726} & 97.652            & 96.402            & 96.248               & 99.399               \\                                                                                                                       & \multicolumn{1}{c|}{\textit{Average}} & 98.880            & 98.195            & 98.055               & \multicolumn{1}{c|}{99.728} & 97.274            & 96.069            & 96.090               & 99.304               \\ \hline
\end{tabular}\label{comparison_table}
\end{table*}

\section{Results}\label{results}
\subsection{Experimental setup}

\subsubsection{Implementation details}

In this research, we used five-fold cross-validation to evaluate the proposed segmentation network, which was trained with a batch size of 4 on an NVIDIA Tesla P100 PCIe GPU. We utilized the Adam optimizer and the initial learning rate was 0.0002. The networks were trained for up to 200 epochs on the Chapman dataset and 100 epochs on the Ningbo dataset. After the first 60 epochs, we applied a Cosine Annealing learning rate scheduler with an initial learning rate of 0.0001 for 20 epochs. Subsequently, the initial learning rate of the Cosine Annealing scheduler was halved every 40 epochs. We also experimented with various ECG data augmentation techniques to balance the class samples; however, we did not observe any significant impact from these augmentations in our experiments, consequently, we chose not to include them in our final approach.

\subsubsection{Evaluation metrics}
To evaluate the proposed method and conduct a comparative analysis with the existing literature, we utilized several key performance metrics: Accuracy, F1-score, Sensitivity, and Specificity. Each metric offers distinct insights into the model's effectiveness. Accuracy quantifies the overall correctness of the model in classifying instances. It is defined as
\begin{equation} \text{Accuracy} = \frac{TP + TN}{TP + TN + FP + FN} \label{eq110
} \end{equation}
where, TP represents the number of correctly identified positive cases, TN indicates the number of correctly identified negative cases, FP is the number of negative instances incorrectly classified as positive, and FN is the number of positive instances incorrectly classified as negative. The F1-score serves as a harmonic mean of Precision and Recall, providing a balanced measure of a model's performance, particularly useful in scenarios involving class imbalance. It is calculated as
\begin{equation} F1\text{-}\text{score} = 2 \times \frac{\text{Precision} \times \text{Recall}}{\text{Precision} + \text{Recall}} =  \frac{2 \cdot \text{TP}}{2 \cdot \text{TP} + \text{FP} + \text{FN}} \label{eq111} 
\end{equation}
Sensitivity, also referred to as Recall or True Positive Rate, assesses the model's ability to correctly identify positive instances. It is defined as
\begin{equation} \text{Sensitivity} = \frac{TP}{TP + FN} \label{eq112
} \end{equation}
Specificity, or True Negative Rate, measures the model's capacity to accurately identify negative instances. It is computed as
\begin{equation} \text{Specificity} = \frac{TN}{TN + FP} \label{eq113
} \end{equation}

By employing these metrics, we gained a comprehensive understanding of the model's performance and relative efficacy compared to other published works. Each metric highlights different aspects of performance, enabling us to identify strengths and areas for improvement in our approach.

\begin{figure*}[t]
\centering

\includegraphics[width=\linewidth]{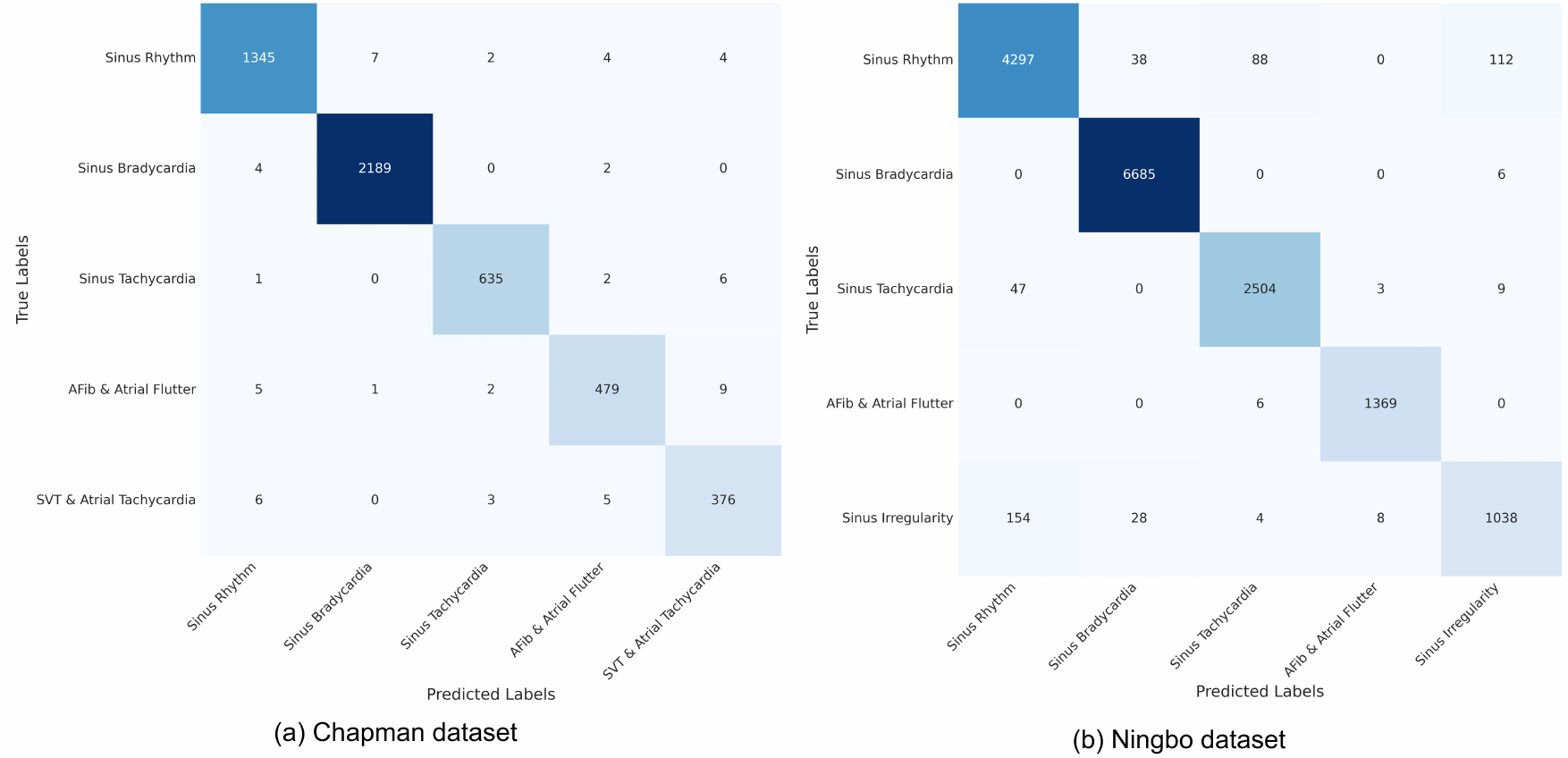}
\caption{Visualization of the aggregated confusion matrix of the proposed network on both datasets.}
\label{fig:confview}
\end{figure*}

\begin{table*}[t]
\caption{Ablation study demonstrating the effect of the newly introduced components in the proposed method.}
\label{ablation_experiment}
\centering
\small
\setlength\tabcolsep{0.4pt}
\begin{tabular}{lcccccccccccc}
\hline
\multicolumn{1}{c}{\multirow{2}{*}{\textbf{Methods}}} & \multicolumn{4}{c}{\textbf{Loss}}                                                                                       & \multicolumn{4}{c}{\textbf{Chapman}}                                                       & \multicolumn{4}{c}{\textbf{Ningbo}}                                                 \\
\multicolumn{1}{c}{}                                  & \textit{$L_{CE}^{b.}$} & \textit{$L_{NCC}^{b.}$} & \textit{$L_{CE}^{f.}$} & \textit{$L_{CE}^{b.+f.}$}                   & \textit{Accuracy} & \textit{F1-score} & \textit{Sensitivity} & \textit{Specificity}        & \textit{Accuracy} & \textit{F1-score} & \textit{Sensitivity} & \textit{Specificity} \\
&                        &                         &                        &                                             &                   &                   &                      &                             &                   &                   &                      &                      \\ \hline
&                        &                         &                        & \multicolumn{1}{c|}{}                       &                   &                   &                      & \multicolumn{1}{c|}{}       &                   &                   &                      &                      \\
(a) Baseline                                          & \ding{51} &                         &                        & \multicolumn{1}{c|}{}                       & 97.857            & 96.286            & 96.299               & \multicolumn{1}{c|}{99.473} & 95.822            & 94.298            & 93.563               & 98.882               \\
&                        &                         &                        & \multicolumn{1}{c|}{}                       &                   &                   &                      & \multicolumn{1}{c|}{}       &                   &                   &                      &                      \\
(b) a + Multiresolution                               & \ding{51} &                         &                        & \multicolumn{1}{c|}{}                       & 98.525            & 97.421            & 97.354               & \multicolumn{1}{c|}{99.631} & 96.584            & 95.136            & 94.270               & 99.083               \\
&                        &                         &                        & \multicolumn{1}{c|}{}                       &                   &                   &                      & \multicolumn{1}{c|}{}       &                   &                   &                      &                      \\
(c) b + XAI guidance                                  & \ding{51} & \ding{51}  &                        & \multicolumn{1}{c|}{}                       & 98.644            & 97.699            & 97.665               & \multicolumn{1}{c|}{99.661} & 96.859            & 95.362            & 95.003               & 99.177               \\
 &                        &                         &                        & \multicolumn{1}{c|}{}                       &                   &                   &                      & \multicolumn{1}{c|}{}       &                   &                   &                      &                      \\
(d) b + Quantitative features                                   & \ding{51} &                         & \ding{51} & \multicolumn{1}{c|}{\ding{51}}                       & 98.585            & 97.528            & 97.527               & \multicolumn{1}{c|}{99.648} & 96.828            & 95.234            & 95.009               & 99.185               \\
 &                        &                         &                        & \multicolumn{1}{c|}{}                       &                   &                   &                      & \multicolumn{1}{c|}{}       &                   &                   &                      &                      \\
(e) b + c + d (EXGnet)                                & \ding{51} & \ding{51}  & \ding{51} & \multicolumn{1}{c|}{\ding{51}} & 98.762            & 97.910            & 97.850               & \multicolumn{1}{c|}{99.698} & 96.932            & 95.527            & 95.403               & 99.210               \\
&                        &                         &                        & \multicolumn{1}{c|}{}                       &                   &                   &                      & \multicolumn{1}{c|}{}       &                   &                   &                      &                      \\
(f) b + c + d (EXGnet-L)                              & \ding{51} & \ding{51}  & \ding{51} & \multicolumn{1}{c|}{\ding{51}} & 98.880            & 98.195            & 98.055               & \multicolumn{1}{c|}{99.728} & 97.274            & 96.069            & 96.090               & 99.304               \\
&                        &                         &                        & \multicolumn{1}{c|}{}                       &                   &                   &                      & \multicolumn{1}{c|}{}       &                   &                   &                      &                      \\ \hline
\end{tabular}
\end{table*}

\subsection{Performance comparison}\label{sec2}

Table \ref{comparison_table} provides a comprehensive comparison of the proposed networks against the most recent methods on the Chapman and Ningbo datasets. In our proposed method, we can either derive predictions from the base model's output, which requires no quantitative features during testing, or utilize the combined model's output when quantitative features are incorporated during inference. To clarify, we refer to the model that uses the combined output as EXGnet-L. Since our primary goal is to avoid using quantitative features during testing, we will concentrate on the output of the base EXGnet.

We employed five-fold cross-validation to evaluate each model, ensuring a thorough and equitable assessment of performance across several metrics, including sensitivity, specificity, accuracy, and F1-score. On the Chapman dataset, EXGnet achieves an impressive average accuracy of 98.762\% and an F1-score of 97.910\%, surpassing all previously mentioned state-of-the-art models. It consistently performs well across all metrics in each of the five folds, attaining the highest average sensitivity and specificity 97.850\% and 99.698\%, respectively. Furthermore, when quantitative features are provided during inference, performance improves significantly, with increases of +0.118\% in accuracy and +0.285\% in F1-score. Among the comparing methods, xECGArch yields the best average accuracy of 98.565\% and an F1-score of 97.451\%. Notably, xECGArch performs better than EXGnet in fold 3, although our method outperforms it in the other folds. When utilizing the combined model with quantitative features, EXGnet also surpasses xECGArch in fold 3. Additionally, G2-ResNeXt demonstrates compelling performance in fold 4 of the Chapman dataset.

In the Ningbo dataset, there is a slight decline in performance across all models, primarily due to the noisy and more challenging ECG signals. However, our proposed EXGnet consistently outperforms other existing methods across all metrics. As shown in Table \ref{comparison_table}, EXGnet achieves 96.932\% accuracy (+0.250\% compared to the second-best model), 95.527\% F1-score (+0.546\% compared to the second-best model), 95.403\% sensitivity (+0.596\% compared to the second-best model), and 99.210\% specificity (+0.083\% compared to the second-best model). Moreover, when quantitative features are provided during testing (EXGnet-L), performance improves significantly (p $<$ 0.001), achieving an average accuracy of 97.274\% and an F1-score of 96.069\%. Among the compared methods, G2-ResNeXt achieves the highest average accuracy at 96.682\%, while xECGArch achieves the best F1-score at 94.981\%. It is evident from Table \ref{comparison_table} that the performance of the Interpretable Multichannel Deep Learning Model is inferior to that of other methods. This may be because this method was designed for classifying twelve-lead ECG signals, and since we applied it to single-lead ECG data, the model may not have adapted well to deliver optimal performance. Overall, the results in Table \ref{comparison_table} highlight how effective our proposed models are for single lead ECG arrhythmia classification. Our network consistently achieves better results in key metrics compared to the existing methods. The compelling performance of EXGnet and EXGnet-L proves they can adapt well to different datasets, making them reliable choices for clinical ECG analysis and setting a new benchmark in the field.

To further observe the classification performance, Figure \ref{fig:confview} shows the aggregated confusion matrices for the EXGnet model on both the Chapman and Ningbo datasets. For the Chapman dataset (Figure \ref{fig:confview}a), the confusion matrix shows excellent performance, with high values concentrated along the main diagonal, indicating a high rate of correct classifications for all five arrhythmia types. Misclassifications are minimal, demonstrating the model's robustness. The confusion matrix for the Ningbo dataset (Figure \ref{fig:confview}b) also shows strong results, though it reflects the more challenging nature of this data. The most prominent confusion is observed between Sinus Rhythm and Sinus Irregularity, with 154 Sinus Irregularity samples misclassified as Sinus Rhythm and 112 Sinus Rhythm signals misclassified as Sinus Irregularity. Despite these challenges, the model maintains compelling accuracy across most classes, reinforcing the overall superior performance reported in Table \ref{comparison_table}.

\begin{figure}[t]
\centering
\includegraphics[width=0.78\linewidth]{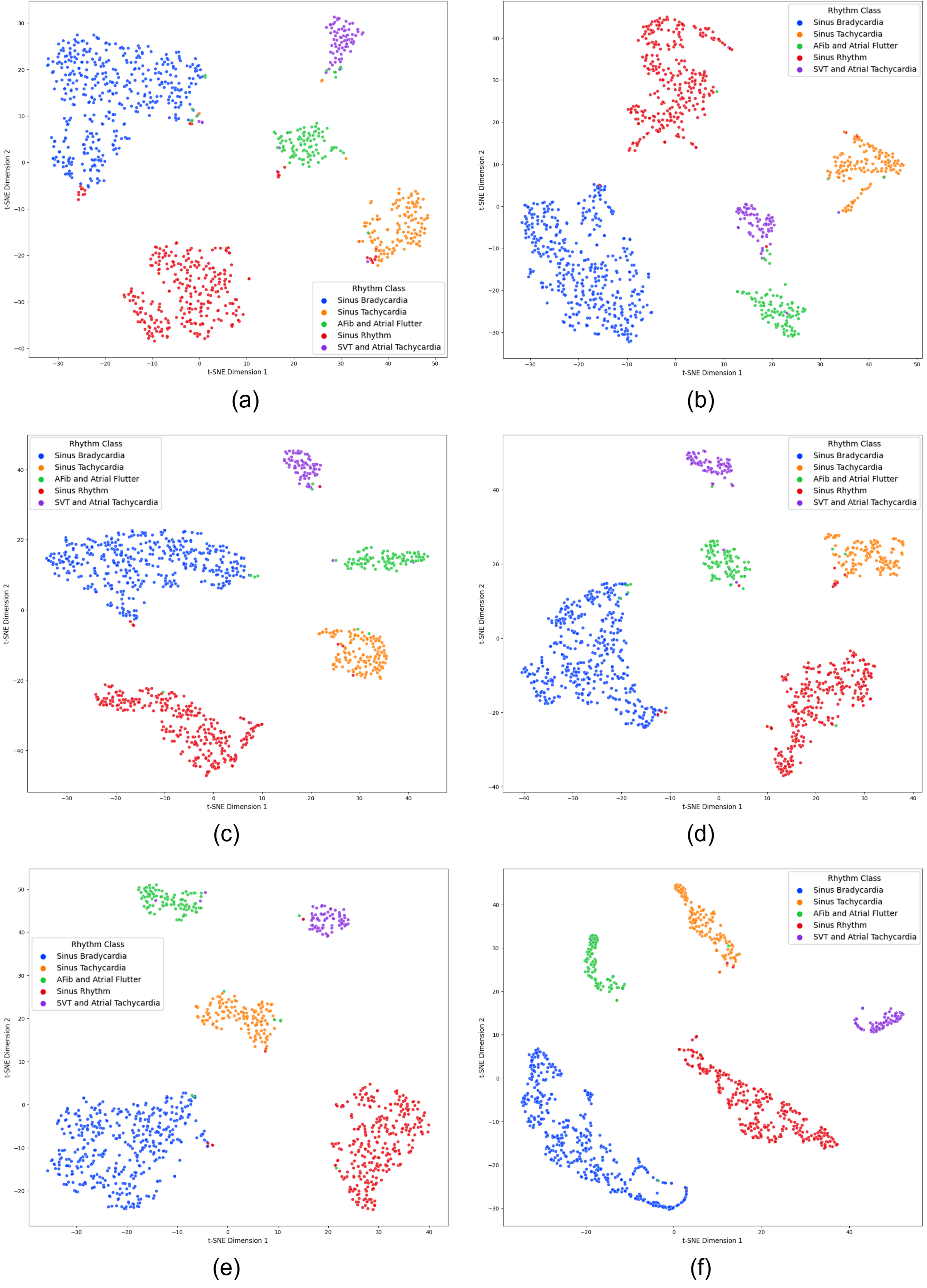}
\caption{t-SNE visualization of the feature maps extracted from the immediate last layer in different ablation study cases for Fold 1 of the Chapman dataset. The cases are: (a) Baseline, (b) a + Multiresolution feature extraction, (c) b + XAI guidance, (d) b + Quantitative features, (e) b + c + d (EXGnet), and (f) b + c + d (EXGnet-L).}
\label{fig:tsneview}
\end{figure}

\begin{figure}[t]
\centering

\includegraphics[width=\linewidth]{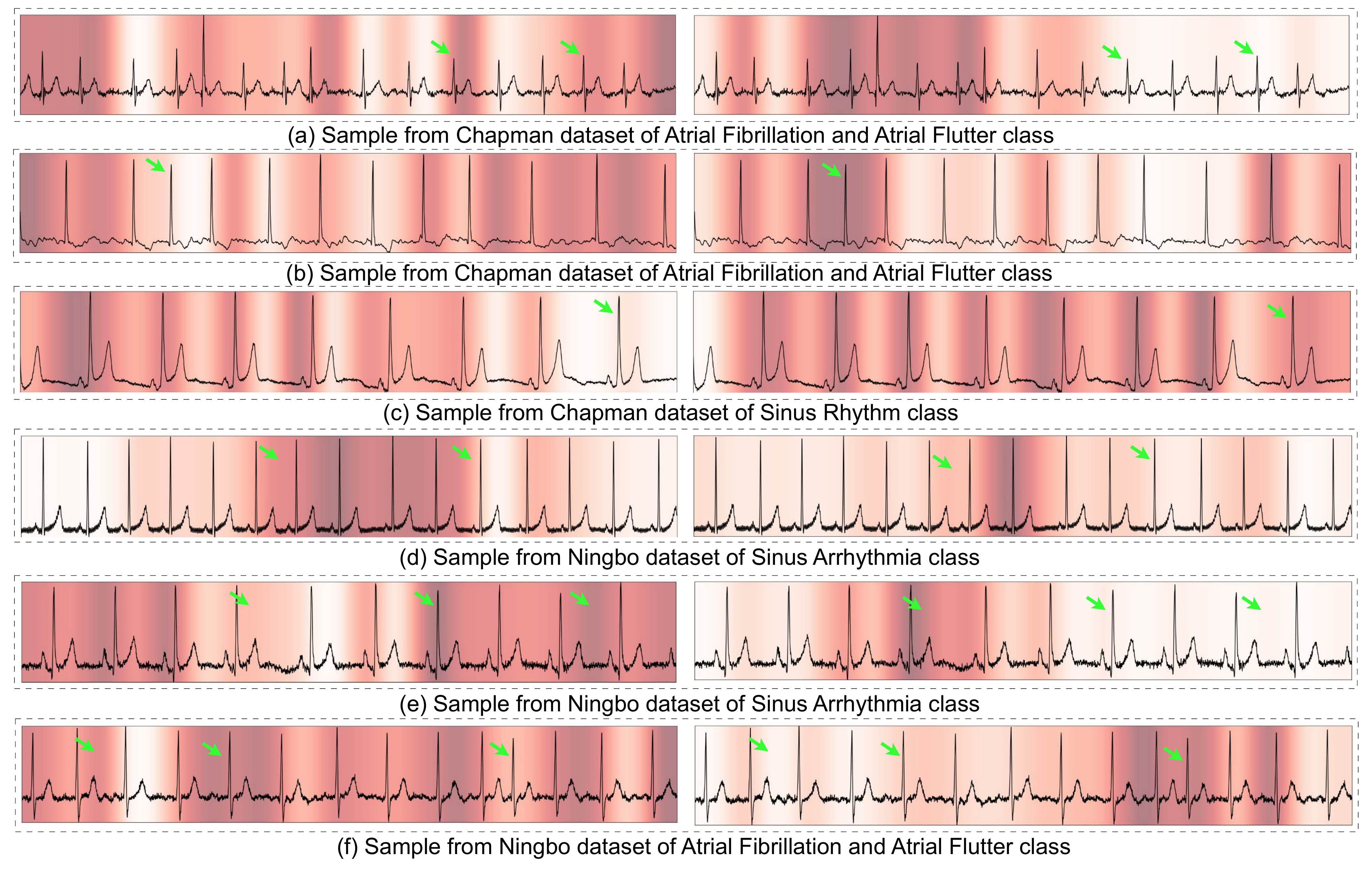}
\caption{Visualization of the effect of XAI guidance on the proposed network. The left side shows the CAM outputs from the model without XAI guidance, while the right side displays the outputs when XAI guidance is applied. Green marks indicate the regions where XAI guidance enhances focus and thus improves reliability.}
\label{fig:xaiview}
\end{figure}

\section{Discussion}\label{discussion}

\subsection{Ablation study}
We conducted a comprehensive ablation study to quantitatively assess the impact of each new component in our proposed method. Table \ref{ablation_experiment} outlines the incremental performance improvements achieved through various configurations: (a) Baseline, (b) Baseline with multiresolution feature extraction, (c) Baseline + multiresolution feature extraction + XAI guidance, (d) Baseline + multiresolution feature extraction + quantitative features integration, (e) Baseline + multiresolution feature extraction + XAI guidance + quantitative features integration (results from the base network EXGnet), and (f) Baseline + multiresolution feature extraction + XAI guidance + quantitative features integration (results from the combined network EXGnet-L). We present the ablation study on the both Chapman and Ningbo datasets, with average five fold results evaluated using Accuracy, F1-score, Sensitivity, and Specificity, as shown in Table \ref{ablation_experiment}.

As shown in Table \ref{ablation_experiment}, the baseline model demonstrated significant performance improvements (p-value $<$ 0.001) with the addition of the multiresolution convolution, yielding increases of 0.668\% and 0.762\% in accuracy, along with 1.135\% and 0.838\% in F1-score for the Chapman and Ningbo datasets, respectively. Introducing XAI guidance further enhanced performance, with accuracy and F1-score gains of 0.119\% and 0.278\% on the Chapman dataset, and 0.275\% and 0.226\% on the Ningbo dataset. Additionally, incorporating quantitative features alongside the multiresolution block also led to improvements in both datasets (+0.06\% and +0.244\% in accuracy, and +0.107\% and +0.098\% in F1-score).

The results clearly indicate that the addition of XAI guidance has a more substantial quantitative impact than integrating quantitative features. However, when both techniques are applied, the accuracy reaches 98.762\% and 96.932\%, with F1-scores of 97.910\% and 95.527\% for the Chapman and Ningbo datasets, respectively, without needing quantitative features during testing. Furthermore, when quantitative features are utilized during testing, the combined model (EXGnet-L) outperformed all metrics across both datasets, with increases of +0.118\% and +0.342\% in accuracy, and +0.285\% and +0.542\% in F1-score. Notably, even with only XAI guidance or quantitative features integration, our proposed network outperformed the existing state-of-the-art models on both dataset.

To provide a qualitative counterpart to our quantitative results, we visualized the high-dimensional feature space using t-SNE. Figure \ref{fig:tsneview} illustrates the resulting feature distributions for each model configuration on fold 1 of the Chapman dataset. Figure \ref{fig:tsneview}(a) shows the feature distribution for the baseline model, where the clusters for different rhythm classes, especially Sinus Rhythm (red) have significant overlap with other classes. With the addition of the multiresolution feature extraction block in Figure \ref{fig:tsneview}(b), the clusters begin to show better separation and become more compact, visually confirming the performance gains reported in Table \ref{ablation_experiment}. The addition of XAI guidance (Figure \ref{fig:tsneview}(c)) and quantitative features (Figure \ref{fig:tsneview}(d)) further improves these clusters, making them tighter and reducing overlapping cases.

A clear improvement is visible in Figure \ref{fig:tsneview}(e), which corresponds to our proposed EXGnet model. Here, the feature clusters are much more distinct and well-defined, showing that the network learned features that can better distinguish between the classes. Finally, Figure \ref{fig:tsneview}(f) shows the feature space for the complete EXGnet-L model. The clusters are extremely tight and clearly separated, with large margins between different classes. This visual evidence strongly supports the numerical results, showing that each proposed component helps in creating a more separable feature representation, which leads to the superior classification performance of the final model.


\begin{table}[h]
\centering
\small
\caption{Quantitative analysis of the interpretability of the proposed network.}
\begin{tabular}{l l c c}
\hline
Dataset & Metric & Without XAI & With XAI \\
\hline

\multirow{3}{*}{Chapman} 
       & NCC & $0.0181 \pm 0.0056$ & $0.6844 \pm 0.0125$ \\
       & L1  & $0.7523 \pm 0.0174$ & $0.2154 \pm 0.0076$ \\
       & L2  & $0.8008 \pm 0.0110$ & $0.3472 \pm 0.0115$ \\
\hline

\multirow{3}{*}{Ningbo} 
       & NCC & $0.0813 \pm 0.0072$ & $0.9192 \pm 0.0454$ \\
       & L1  & $0.5662 \pm 0.0307$ & $0.1411 \pm 0.0106$ \\
       & L2  & $0.6368 \pm 0.0355$ & $0.2915 \pm 0.0211$ \\
\hline

\end{tabular}

\label{tab:xai_comparison}
\end{table}

\begin{figure*}[t]
\centering
\includegraphics[width=\linewidth]{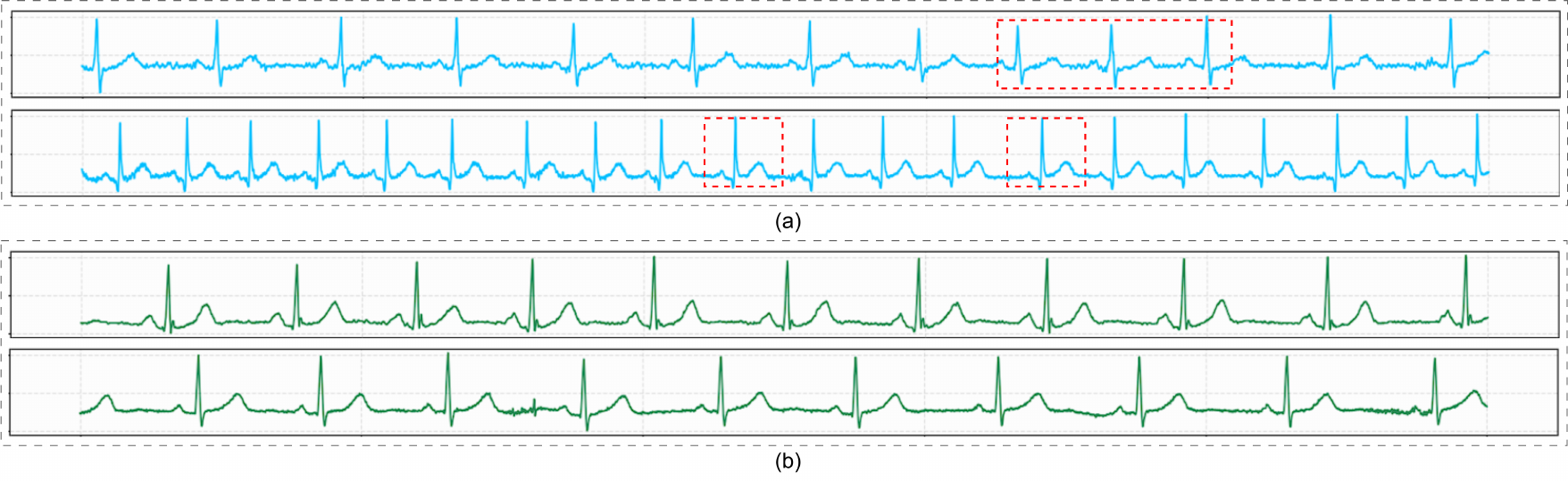}
\caption{Potential anomalous cases identified from the Ningbo dataset. (a) Real label is Sinus Rhythm, but predicted as Sinus Irregularity. (b) Real label is Sinus Irregularity, but predicted as Sinus Rhythm.}
\label{fig:misningview}
\end{figure*}

\begin{figure}[t]
\centering

\includegraphics[width=0.6\linewidth]{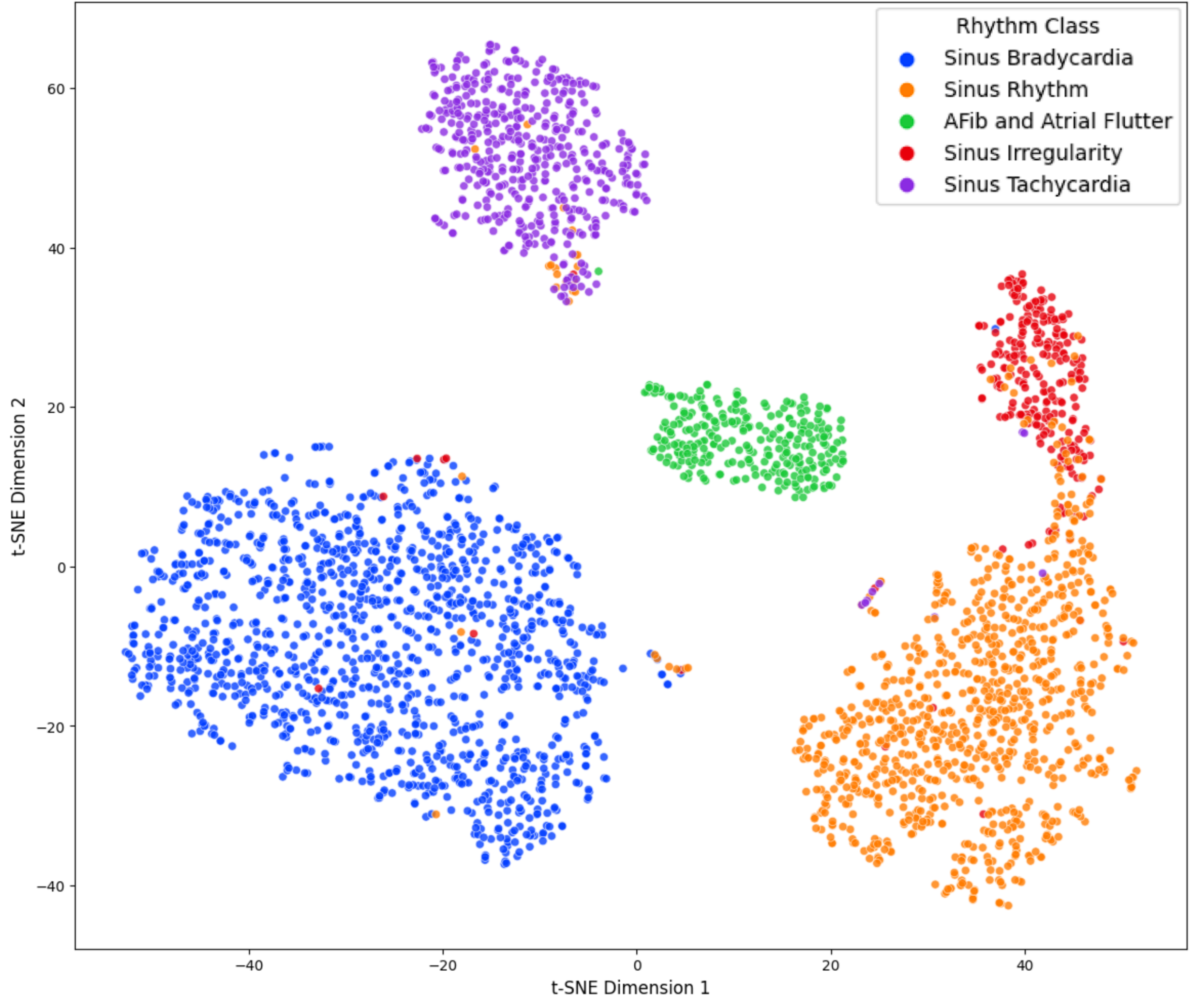}
\caption{t-SNE visualization of the feature maps extracted from the immediate last layer for Fold 5 of the Ningbo dataset.}
\label{fig:tsneningview}
\end{figure}

\subsection{Evaluation of trustworthiness}

In the ablation study section, we showed the quantitative impact of incorporating XAI guidance (see Table \ref{ablation_experiment}). However, these quantitative results alone do not reveal whether the network’s correct predictions are based on the specific segments of the ECG signal that a cardiologist would typically analyze during detecting arrhythmia. It is possible for a network to make correct predictions based on irrelevant regions of the ECG signals. Therefore, it is important to assess the focusing areas in the ECG signal by the network. To evaluate the trustworthiness of our proposed network, we used Grad-CAM to determine which parts of the signal the network focuses on.

Figure \ref{fig:xaiview} illustrates the effectiveness of XAI guidance. The left side shows CAMs without XAI guidance, while the right side presents CAMs with XAI guidance applied. In sample (a), the left CAM highlights some incorrect regions that contributed to the prediction. In contrast, the right CAM, utilizing XAI guidance, displays improved focus and reliability. Although the network correctly identified the arrhythmia in both cases, XAI guidance enhanced the trustworthiness of the results.

A similar scenario is observed in sample (e), where the left CAM indicates that the network did not focus on the most critical features of the ECG signal. In contrast, with XAI guidance, the model learns to concentrate on the relevant areas that cardiologists typically consider when making decisions. Thus, Figure \ref{fig:xaiview} clearly illustrates the qualitative impact of XAI guidance and underscores its importance in achieving reliable predictions. To quantitatively assess the trustworthiness of the proposed technique, we introduced an innovative evaluation approach. Specifically, the CAMs generated by Grad-CAM are compared with ground truth CAMs using three metrics: NCC, L1 distance, and L2 distance. As shown in Table~\ref{tab:xai_comparison}, the inclusion of XAI guidance during training significantly improves the alignment between predicted and true CAMs across both datasets. Specifically, the NCC scores increase substantially, indicating a higher similarity, while the L1 and L2 errors decrease, reflecting better localization accuracy and reduced deviation. These results demonstrate the effectiveness of the proposed XAI-guided model in producing more trustworthy and interpretable explanations.

\subsection{Confusing cases identified in the Ningbo dataset}
Figure~\ref{fig:confview}(b) reveals that a significant portion of signals from Ningbo dataset are confused between Sinus Rhythm and Sinus Irregularity. To further explore this observation, we examined the t-SNE visualization in Figure~\ref{fig:tsneningview}, based on fold 5 of the Ningbo dataset. This plot reflects how the network separates different classes in the high-dimensional feature space. Notably, there is a considerable overlap between the Sinus Rhythm and Sinus Irregularity classes, suggesting that the network perceives similar features in some instances of both categories. While it is true that these cases are inherently challenging for the model, further inspection revealed the presence of certain anomalies that could be contributing to this confusion. We observed that a noticeable number of signals labeled as Sinus Rhythm exhibit features commonly associated with Sinus Irregularity, and vice versa. Representative examples from the validation set of fold 5 are shown in Figure~\ref{fig:misningview}.

Figure~\ref{fig:misningview}(a) presents two signals with ground truth labels of Sinus Rhythm. However, upon closer observation, subtle regions of irregular rhythm are visible, which likely led the network to predict them as Sinus Irregularity. In contrast, Figure~\ref{fig:misningview}(b) displays two signals labeled as Sinus Irregularity that were predicted as Sinus Rhythm. Interestingly, these signals show only minimal rhythm irregularities, even less pronounced than those in Figure~\ref{fig:misningview}(a). Such borderline cases challenge the model’s ability to learn class-distinctive features effectively.

It is important to note that an ideal Sinus Irregular Rhythm should display a noticeable variation in heart rhythm. Another type of anomaly found in the dataset involves incorrect labeling between Sinus Bradycardia and Sinus Rhythm, as well as between Sinus Tachycardia and Sinus Rhythm. For example, the second signal in Figure~\ref{fig:misningview}(a) appears to have a heart rate of approximately 120 beats per minute, yet it is labeled as Sinus Rhythm instead of Sinus Tachycardia. These mislabeled or ambiguous cases can significantly mislead the network during training. Although limitations exist within the model itself, achieving higher performance and reliability ultimately depends on the accuracy and consistency of the dataset labels.

\subsection{Limitations and future works}
Our proposed network shows reliable and promising results when compared with state-of-the-art approaches. It is not without limitations, though. First, the number of quantitative features trained could be expanded to allow the model to learn more statistical characteristics. Second, we relied solely on a Grad-CAM inspired CAM generation technique for XAI guidance due to its simplicity. While effective, other advanced CAM-generating techniques might enhance the network's ability to focus on critical features of the ECG signal.

This work serves as an initial exploration of the positive effects of XAI guidance in ECG arrythmia classification. Furthermore, we limited our ground truth generation for XAI guidance to RR intervals variability for the sake of simplicity and automation. However, complex arrhythmias also involve other important features, such as P waves absence, ST elevation, and T waves position. Including these would require manual identification of the affected regions, necessitating collaboration with expert cardiologists and additional labor time. Thus, addressing these limitations will be a central focus of our future work.

\section{Conclusion}\label{conclusion}

In this work, we introduced EXGnet, a reliable and trustworthy deep learning network developed for single-lead ECG arrhythmia classification, with a strong focus on enhancing interpretability through XAI guidance during training. By supervising the learning process with ground truth CAMs generated through an innovative heart rate variability based technique, the model is guided to focus on clinically meaningful regions of the ECG signal. This guidance not only improves classification performance but also significantly enhances the trustworthiness and transparency of the model's decision-making process.
To enrich the model's learning with multi-domain knowledge, we incorporated quantitative features during training. These train-only features provide the model with complementary information beyond what is learned from raw data. While these features are not required during testing, their optional inclusion at inference leads to notable performance improvements.
Furthermore, we designed a multiresolution feature extraction strategy to capture relevant patterns at different temporal scales. This design choice enables the model to maintain high performance while keeping computational demands low, making it well suited for real-time applications in edge environments.
Our extensive experiments demonstrate that EXGnet surpasses recent state-of-the-art models across all key metrics. These results highlight the potential of EXGnet for practical deployment in portable and edge-based single-lead ECG monitoring systems, including smartphones and wearable healthcare devices.

\section*{Acknowledgements}
We would like to express our gratitude to Teton Private Ltd. for their partial financial support in conducting this project.

\section*{CRediT authorship contribution statement}
Tushar Talukder Showrav: Original draft preparation, Writing – review \& editing, Conceptualization, Methodology, Software, Validation, Investigation, Visualization. Soyabul Islam Lincoln: Original draft preparation, Software, Validation, Investigation. Md. Kamrul Hasan: Conceptualization, Supervision, Writing – review \& editing.

\section*{Data Availability Statement}
No new data were generated or analyzed during this study.

\section*{Conflict of Interest}
The authors state that they have no conflicts of interest.

\bibliographystyle{unsrtnat}
\bibliography{references}  






\end{document}